\def\year{2021}\relax
\documentclass[letterpaper]{article} 
\usepackage{aaai21}  
\usepackage{times}  
\usepackage{helvet} 
\usepackage{courier}  
\usepackage[hyphens]{url}  
\usepackage{graphicx} 
\usepackage{natbib}  
\usepackage{caption} 
\usepackage{amssymb}
\usepackage[ruled,vlined,linesnumbered]{algorithm2e}
\usepackage{comment}

\newtheorem{definition}{Definition}
\newtheorem{proposition}{Proposition}
\newtheorem{corollary}{Corollary}

\title{Multi-Goal Multi-Agent Path Finding via\\Decoupled and Integrated Goal Vertex Ordering}
\author {
    Pavel Surynek \thanks{The author has been supported by GA\v{C}R - the Czech Science Foundation, grant registration number 19-17966S.}\\
}

\affiliations{
    Czech Technical University in Prague\\
    Faculty of Information Technology\\ 
    Tha\'{a}kurova 9, 160 00 Praha 6, Czechia\\
    pavel.surynek@fit.cvut.cz
}

\begin{document}

\maketitle

\begin{abstract}
We introduce multi-goal multi agent path finding (MAPF$^{MG}$) which generalizes the standard discrete multi-agent path finding (MAPF) problem. While the task in MAPF is to navigate agents in an undirected graph from their starting vertices to one individual goal vertex per agent, MAPF$^{MG}$ assigns each agent multiple goal vertices and the task is to visit each of them at least once. Solving MAPF$^{MG}$ not only requires finding collision free paths for individual agents but also determining the order of visiting agent's goal vertices so that common objectives like the sum-of-costs are optimized. We suggest two novel algorithms using different paradigms to address  MAPF$^{MG}$: a heuristic search-based search algorithm called Hamiltonian-CBS (HCBS) and a compilation-based algorithm built using the SMT paradigm, called SMT-Hamiltonian-CBS (SMT-HCBS). Experimental comparison suggests limitations of compilation-based approach.
\end{abstract}

\section{Introduction}
\noindent
Mutli-agent path finding (MAPF) \cite{DBLP:conf/aiide/Silver05,DBLP:journals/jair/Ryan08,DBLP:conf/icra/Surynek09,DBLP:conf/ijcai/LunaB11,DBLP:journals/jair/WangB11,DBLP:journals/aimatters/MaK17} is an abstraction for many real-life problems where agents, both autonomous or passive, need to be moved (see \cite{DBLP:conf/socs/FelnerSSBGSSWS17,DBLP:journals/aimatters/MaK17} for a survey). The environment in MAPF is modeled as an undirected graph where vertices represent positions and edges define the topology \footnote{Almost identical setup is used in {\em graph pebbling} \cite{DBLP:journals/jasss/KornhauserWR09,DBLP:journals/algorithms/Parberry15,DBLP:journals/jsc/RatnerW90} where however the focus is rather on computational complexity issues and theory.}.

The standard variant of MAPF assumes that each agent starts in a given starting vertex and its task is to reach unique individual goal vertex. While such formalization encompass many real-life navigation tasks \cite{DBLP:conf/atal/CapNVP13,DBLP:journals/corr/0001KA0HKUXTS17} there still exist problems especially in logistic domain where the standard MAPF lacks expressiveness.

Such problems that cannot be expressed using the standard MAPF include situations where agents have multiple goals so that instead of reaching single goal location agents need to perform a round-trip to service a set of goals. Many real-life applications requires that agents perform certain task at each of multiple goal locations such as performing maintenance or pickup operation \cite{DBLP:journals/cor/PansartCC18,DBLP:journals/eor/BriantCCCLO20}.

Having multiple goals per agent adds a significant new challenge to the problem consisting of determining the order of visiting agent's goal vertices. Hence the ordering of goals as well as non-conflicting path finding are subject to decision making which in addition to this aims on optimization of various objectives such as commonly adopted sum-of-costs \cite{DBLP:conf/aaai/Standley10,DBLP:journals/ai/SharonSGF13}

We introduce a problem we call a {\em multi-goal multi-agent path finding} (MAPF$^{MG}$). The problem shares movement rules with the standard MAPF but generalizes it by allowing each agent to have multiple goal vertices. Each of the agent's goal vertices must be visited at least once to successfully solve MAPF$^{MG}$.

\subsection{Contribution}
\noindent

We introduce two novel solving algorithms for MAPF$^{MG}$: a search-based Hamiltonian CBS (HCBS), a derivative of the CBS algorithm \cite{DBLP:journals/ai/SharonSFS15}, and a compilation-based algorithm, called SMT-Hamiltonian-CBS, derived from SMT-CBS \cite{DBLP:conf/ijcai/Surynek19}.

The important feature of HCBS is that it decouples goal vertex ordering from collision free path finding making HCBS a three level search algorithm where at the high-level, the standard CBS conflict resolution search is performed and at the low level search for collision-free (Hamiltonian) path is done further divided into two levels. The higher level of path-finding determines the goal vertex ordering while the lower level finds collision free path visiting all goals following the order determined by the higher level.

The compilation-based approach uses ideas from solvers for {\em satisfiability modulo theories} (SMT) \cite{DBLP:conf/cav/BarrettMS05,DBLP:series/faia/BarrettSST09} namely lazy construction of formulae in the target formalism. The significant feature of this approach is that collision avoidance and goal vertex ordering are integrated and solved at once.

The paper is organized as follows: basic definitions from MAPF and its properties are introduced first. Then search-based HCBS and related MAPF$^{MG}$ specific heuristics are developed. Compilation-based SMT-HCBS follows in the next section. Finally both new algorithms are experimentally evaluated and compared.

\section{Background and Definitions}
\noindent
We first recall the concepts from the standard multi-agent path finding (MAPF) that we inherit into MAPF$^{MG}$. Agents in MAPF are placed in vertices of an undirected graph so that there is at most one agent per vertex. Formally, $s: A \mapsto V$ is a {\em configuration} of agents in vertices of the graph. A configuration can be transformed instantaneously to the next one by {\em valid} movements of agents; the next configuration corresponds to the next {\em time step}. An agent can move into another vertex across an edge provided the target vertex is vacant or being simultaneously vacated by another agent. A {\em collision} occurs if two agents appear in the same vertex ({\em vertex collision}) or if two agents simultaneously cross the same edge in opposite directions ({\em edge collision}) \cite{DBLP:journals/ai/SharonSFS15}. \footnote{Different variants of valid movements exist such a permitting agents to move into vacant vertices only \cite{DBLP:journals/ci/Surynek14}.} The configuration at time step $t$ is denoted $s_t$.

\begin{definition}
{\em Mutli-goal multi-agent path finding} (MAPF$^{MG}$) is a 4-tuple $\Theta = (G=(V,E), A, s_0, g)$ where $G=(V,E)$ is an undirected graph, $A=\{a_1, a_2, ...,a_k\}$ with $k \in \mathbb{N}$ is a set of agents where $k \leq |V|$, $s: A \mapsto V$ represents agents' starting vertices (starting configuration), and $g: A \mapsto 2^V$ assigns a set of goal vertices to each agent.
\end{definition}

\begin{figure}[t]
    \centering
    \includegraphics[trim={3.3cm 23cm 3.3cm 2.5cm},clip,width=0.55\textwidth]{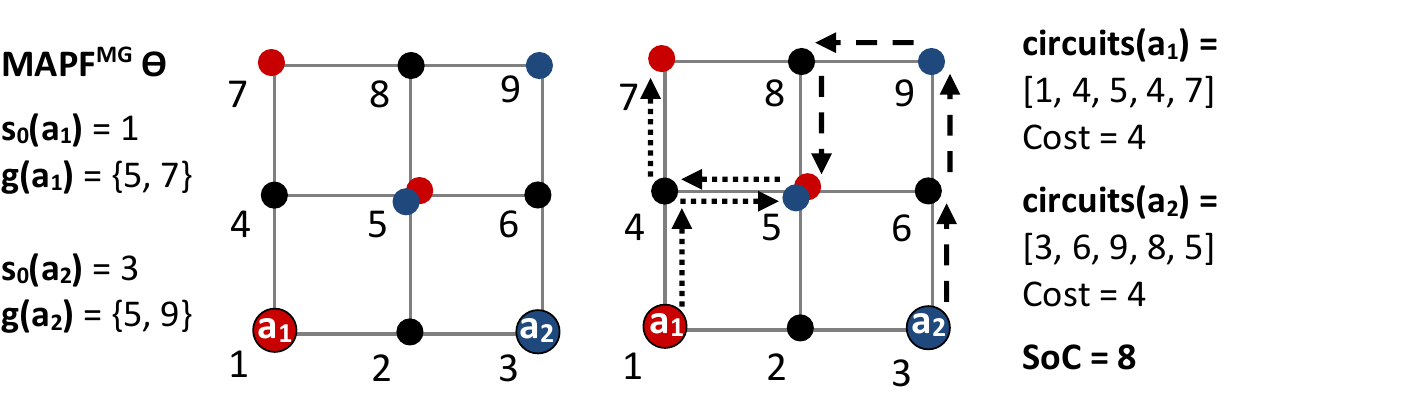}
    \vspace{-0.8cm}\caption{A MAPF$^{MG}$ instance with and its sum-of-costs optimal solution.}
    \label{figure-MAPF-MG}
\end{figure}

Each agent in MAPF$^{MG}$ has the task to visit its goal vertices. Agent's goal vertices can be visited in an arbitrary order but each goal must be visited by the agent at least once. Various objectives can be taken into account. We develop all concepts here for the {\em sum-of-costs} objective commonly adopted in MAPF but different cumulative objectives can be used as well \cite{DBLP:conf/socs/SurynekFSB16}.

Formally, the MAPF$^{MG}$ solution is a sequence of configurations that can be obtained via valid moves from the starting configuration $s_0$ following the standard MAPF movement rules and each agent visits each of its goal vertices at least once:

\begin{definition}
A solution to MAPF$^{MG}$ $\Theta = (G=(V,E), A, s_0, g)$ is a sequence of configurations $S = [s_0,s_1,s_2,...,s_{t_M}]$ such that $s_{t+1}$ results from $s_t$ for all $t \in \{0,1,...{t_M}-1\}$ via valid MAPF movements and $\forall a_i \in A$ it holds that $(\forall v_g \in g(a_i))$ $(\exists t \in \{1,2,...,{t_M}\})$ $(s_t(a_i) = v_g)$.
\end{definition}

Given a solution $[s_0,s_1,s_2,...,s_m]$, $t_m$ is the {\em makespan}, denoted $\mathit{Make}(S)$. We define the {\em sum-of-costs} as the sum of costs of individual agents: $\mathit{SoC}(S) = \sum_{i=1}^{k}{\mathit{Cost}(a_i)}$, where individual cost of agent $a_i$ is defined as: $\mathit{Cost}(a_i) = \min$ $\{ t_c \;|\; (\forall v_g \in g(a_i))$ $(\exists t \in {1,2,...,t_c})$ $(s_t(a_i) = v_g)\}$.


Let us note that the $\mathit{SoC}(S)$ accumulates unit costs of actions including wait actions until all goals are visited. After that we do not care about agent's movements - it can stay in the last visited goal or it can move arbitrarily. This is one possible definition of MAPF$^{MG}$. We can alternatively require that all agents should return to their starting vertices and/or stay in the final goal. The developed solving algorithms require only minor modifications to be applied for any of such MAPF$^{MG}$ variants.

MAPF$^{MG}$ as well as the original MAPF is NP-hard \cite{DBLP:journals/jsc/RatnerW90,DBLP:conf/aaai/Surynek10,DBLP:conf/aaai/YuL13,DBLP:journals/corr/YuL15c}. A trivial difference can be observed for a special case with one agent. Finding a {sum-of-costs} optimal solution to MAPF with one agent is in P as it corresponds to finding a shortest path in $G$. On the other hand sum-of-costs optimal MAPF$^{MG}$ with one agent is still NP-hard as it corresponds to finding Hamiltonian path \cite{DBLP:journals/ipl/RahmanK05} in a graph.

An example of MAPF$^{MG}$ and its sum-of-costs optimal solution is shown in Figure \ref{figure-MAPF-MG}.

\section{Hamiltonian Conflict-based Search: HCBS}
\noindent
We suggest a novel algorithm called Hamiltonian Conflict-based Search (Hamiltonian CBS, HCBS) which shares the high level structure with the CBS algorithm \cite{DBLP:journals/ai/SharonSFS15}, one of the most commonly used algorithm for the standard MAPF \cite{DBLP:conf/aips/LiHS0K19,DBLP:conf/socs/LiBF0K19,DBLP:conf/aips/ZhangLSKK20,DBLP:conf/aips/0001GHS0K20,DBLP:conf/socs/BoyarskiHSBF20}.

When trying to use CBS for MAPF$^{MG}$, the significant challenge is represented by the fact that at the low level there is no longer search for a minimum cost path with respect to the set of conflicts, a polynomial-time problem, but rather the search for a minimum cost Hamiltonian path (which even without a set of conflicts is a difficult problem).

\begin{definition}
A {\em Hamiltonian path} (HP) in $G$ starting at $u \in V$ covering a subset of vertices $U \subseteq V$ is a sequence of vertices denoted $H_P(u,U) = [h_0,h_1,...,h_{t_H}]$ such that $h_1 = u$, $\{h_t,h_{t+1}\} \in E$ for $t \in \{0,1,...,t_H - 1\}$, and for each $v \in U$ $\exists t \in \{0,1,...,t_H\}$ such that $h_t = v$. The cost of Hamiltonian path correspond to the number of its edges: $\mathit{Cost}(H_P(u,U)) = t_H - 1$.
\end{definition}

Such CBS hence consists of potentially exponential-sized conflict resolution at the high-level where each conflict leads to exponential-time search for a fresh Hamiltonian path, resulting in an algorithm with high exponential factor. Despite prohibitive theoretical complexity such algorithm could be feasible provided that low level search is fast enough.

\subsection{Decoupled Goal Ordering}

It turned out that the key to adapting CBS for MAPF$^{MG}$ is to decouple goal vertex ordering from conflict avoidance at the low level. The search for a Hamiltonian path going through agent's goal vertices with respect to a set of conflicts is done in two level fashion. At the higl-level (of this low level) we are trying to determine optimal ordering of agent's goal vertices. To this purpose we made use of the A* algorithm \cite{DBLP:journals/tssc/HartNR68} that searches the space of possible {\bf permutations} of agent's goals. After determining the goal vertex to visit as the next one the algorithm searches for the shortest path observing the conflicts that interconnects the next goal vertex with the current one. The search for the shortest path is done by another instance of A*. Another important factor for the performance of the decoupled approach are the heuristics.

\subsection{Minimum Spanning Tree Heuristic}
We define a variant of {\em spanning tree} with respect to a subset of vertices of undirected graph $G=(V,E)$. 

\begin{definition}
A {\em spanning tree} (ST) of an undirected graph $G=(V,E)$ with respect to a subset of vertices $U \subseteq V$, denoted $T_S(U) = (V_U,E_U)$ is a tree covering $U$, that is, $U \subseteq V_U \subseteq V$ and $E_U \subseteq E$. The {\em cost} of a spanning tree is defined as the number of edges included in the tree: $\mathit{Cost}(T_S(U)) = |E_U|$. A {\em minimum spanning tree} (MST) with respect to $U$ is a spanning tree with minimum cost.
\end{definition}

Observe that a spanning $T_S(U)$ may contain other vertices except those in $U$ to keep it connected. We also use the notation $T_S(u,U)$ for $u \in V$ denoting a spanning tree covering the set of vertices $\{u\} \cup U$.

The important property of MST is that it can be found in polynomial time with respect to $G$ \cite{boruvka1926,DBLP:journals/dm/NesetrilMN01}. We can use MST for determining whether there exists a spanning tree in $G$ of given cost. Another important property is that the cost of MST can serve as the lower bound for the cost of shortest Hamiltonian path:

\begin{proposition}
For any $u \in V$ and $U \subseteq V$ it holds that $min\{\mathit{Cost}(T_S(u,U))\}  \leq min \{\mathit{Cost}(H_P(u,U))\}$.
\label{prop-Hamiltonian}
\end{proposition}

This enables us to use the concept of MST as a basis for {\em consistent} A* heuristic (proof of monotonicity omitted). The HCBS algorithm is described using pseudo-code as Algorithm \ref{alg-HCBS}.

The {\bf high-level} represented by HCBS$\mathit{_{conflicts}}$ follows the standard CBS. It construct a {\em constraint tree} (CT) in breadth-first search manner where each node $N$ contains a set of collision avoidance constraints $N.constraints$ - a set of triples $(a_i,v,t)$ forbidding occurrence of agent $a_i$ at $v \in V$ at time step $t$, a solution $N.circuits$ - a set of $k$ Hamiltonian paths for individual agents covering their individual goals, and the sum-of-costs $N.\mathit{SoC}$ of the current solution. Nodes are stored in a priority queue \textsc{Open} and processed in the ascending order according to $N.\mathit{SoC}$.

Initially the shortest Hamiltonian paths for each agent are determined and corresponding node is stored into \textsc{Open} (lines 2-5). At a general step, HCBS takes a node $N$ with the minimum sum-of-costs from \textsc{Open} and checks whether it represents a valid solution w.r.t. MAPF$^{MG}$ rules (lines 7-9). If there are no collisions, then the valid solution $N.\mathit{circuits}$ is returned and the algorithm finishes (lines 10-11). Otherwise the search branches by creating two new nodes $N_1$ and $N_2$ - successors of $N$. Assuming a collision $(a_i, a_j, v, t)$ between agents $a_i$ and $a_j$ at vertex $v$ at time step $t$, this can be avoided if either $a_i$ or $a_j$ does not reside at $v$ at time step $t$. This requirement correspond to two conflict avoidance constraints in successor nodes $N_1$ and $N_2$ that inherit set of constraints from $N$ as follows: $N_1.\mathit{constraints} = N.\mathit{constraints} \cup \{(a_i,v,t)\}$ and $N_2.\mathit{constraints} = N.\mathit{constraints} \cup \{(a_j,v,t)\}$ (line 14). In addition to this, fresh Hamiltonian path for affected for agents $a_i$ and $a_j$ are recomputed in respective nodes (lines 15-17).

The {\bf low-level} is represented by HCBS$\mathit{_{ordering}}$ that determines the ordering of agent's goal vertices. This is an A*-based search of the space of goal vertex permutations. Each node $N$ of the search tree consists of $N.u$ a current vertex, starting at $s_0(a_i)$, a set of visited goals $N.finished$, an A*'s g-value and h-value: $N.g$ and $N.h$ where $N.g$ corresponds to the actual cost of partially constructed Hamiltonian finishing in $N.u$, and $N.h$ is a lower bound estimation of the cost the remaining part of the Hamiltonian path calculated as the cost of MST starting in $N.u$ and covering the remaining goals. Finally, $N.circuits$ is the partial Hamiltonian path itself.

The very low-level search is done by another instance of A* (line 33) which interconnects the current vertex $N.u$ with the candidate for the next goal $v$ by a shortest path taking into account conflicts $constraints$ from the very high-level CBS-style search. This low level search could be equipped by another 

Altogether HCBS consists of {\bf thee levels of search} in three different spaces: (i) space of conflicts, (ii) goal ordering space, and (iii) path space.

\begin{algorithm}[t!]
\begin{footnotesize}
\SetKwBlock{NRICL}{HCBS$\mathit{_{conflicts}}$ ($\Theta = (G=(V,E),A,s_0,g)$)}{end} \NRICL{
    $N.constraints \gets \emptyset$ \\
    $N.circuits \gets$ $\{circuit^*(a_i)$ a shortest Hamiltonian path from $s_0(a_i)$ covering $g(a_i)\;|\; i = 1,2,...,k\}$\\
    $N.\mathit{SoC} \gets \sum_{i=1}^k{\mathit{Cost}(N.circuits(a_i))}$ \\    
    insert $(\mathit{SoC},N)$ into $\textsc{Open}$ \\
    \While {$\textsc{Open} \neq \emptyset$} {
        $(key,N) \gets$ min-Key($\textsc{Open}$)\\
        remove-Min-Key($\textsc{Open}$)\\
        $collisions \gets$ validate($N.circuits$)\\
        \If {$collisions = \emptyset$}{
            \Return $N.circuits$\\
        }
        let $(a_i,a_j,v,t) \in collisions$\\      
        \For {each $a \in \{a_i,a_j\}$}{
       	$N'.constraints \gets N.constraints \cup \{(a,v,t)\}$\\
        	$N'.circuits \gets N.circuits$\\
        	$N'.circuits(a_i) \gets$ HCBS$\mathit{_{ordering}}$ $(\Theta$, $a$, $N'.constraints)$ \\
             $\mathit{SoC'} \gets \sum_{i=1}^k{\mathit{Cost}(N'.circuits(a_i))}$ \\           	
		insert $(\mathit{SoC'},N')$ into $\textsc{Open}$ \\
        }
     }    
}

\SetKwBlock{NRICL}{HCBS$\mathit{_{ordering}}$ $(\Theta, a_i, constraints)$}{end} \NRICL{
    let $\Theta = (G=(V,E),A,s_0,g)$ \\
    $N.u \gets s_0(a_i)$; $N.finished \gets \emptyset$ \\
    $T_S(N.u,g(a_i)) \gets$ construct-MST$(s_0(a_i), g_R, G)$ \\
    $N.g \gets 0$; $N.h \gets \mathit{Cost}(T_S(N.u,g_R))$ \\
    $N.circuit \gets []$ \\
    insert $(N.g +N.h,N)$ into $\textsc{Open}$ \\ 
    \While {$\textsc{Open} \neq \emptyset$} {
        $(key,N) \gets$ min-Key($\textsc{Open}$)\\
        remove-Min-Key($\textsc{Open}$)\\
        \If {$N.finished = g(a_i)$}{
            \Return $N.circuit$\\
        }
        \Else{
            \For {each $v \in g(a_i) \setminus N.finished$}{
              $path \gets$ A*$(N.u, v, constraints, N.g)$ \\
              \If{$path \neq \mathit{Fail}$}{
                  $N'.u \gets v$ \\
                  $N'.finished \gets N.finished \cup \{v\}$ \\
                  $g'_R \gets g(a_i) \setminus N'.finished$ \\
                  $T_S(N'.u,g'_R) \gets$ construct-MST$(N'.u, g'_R, G)$ \\
                  $N'.g \gets N.g + \mathit{Cost}(path)$ \\
                  $N'.h \gets \mathit{Cost}(T_S(N'.u,g'_R))$ \\
                  $N'h \gets N.circuit \cdot path$ \\
                  insert $(N'.g +N'.h,N)$ into $\textsc{Open}$ \\               
              }
          }
       }
    }
    \Return $\mathit{Fail}$ \\
}
\caption{HCBS algorithm for MAPF$^{MG}$.} \label{alg-HCBS}
\end{footnotesize}
\end{algorithm}

Combining soundness and optimality of CBS with properties of MST-based heuristic for goal vertex ordering we can state the following proposition.

\begin{proposition}
The HCBS algorithm returns sum-of-costs optimal solution for given input MAPF$^{MG}$ $\Theta$.
\label{prop-HCBS}
\end{proposition}

\section{Compilation-based Approach: SMT-HCBS}
\noindent
The second approach for solving MAPF$^{MG}$ is based on reduction of MAPF$^{MG}$ to a series of propositional formulae that are decided by the external SAT solver \cite{DBLP:journals/ijait/AudemardS18}. Our new algorithm called SMT-Hamiltonian-CBS combines existing SMT-CBS \cite{DBLP:conf/ijcai/Surynek19} with MAPF$^{MG}$ specific generation of target propositional formuale.

\subsection{Time Expansion and Decision Diagrams}

Construction of a propositional formula corresponding to solvability of a given MAPF as used in SMT-CBS relies on the time expansion of underlying graph $G$. Having $\mathit{SoC}$, the basic variant of time expansion determines the maximum number of time steps ({\em makespan}) $t_M$ such that every possible solution of with the sum-of-costs less than or equal to $\mathit{SoC}$ fits in $\mathit{Make}$ timesteps.

The time expansion makes copies of vertices $V$ for each timestep $t=0,1,2,...,t_M$. That is, we have vertices $v^t$ for each $v \in V$ and time step $t$. Edges from $G$ are converted to directed edges interconnecting timesteps in the time expansion. Directed edges $(u^t,v^{t+1})$ are introduced for $t=0,1,...,t_M-1$ whenever there is $\{u,v\} \in E$. Wait actions are modeled by introducing edges $(u^t,u^{t+1})$. A directed path in the time expansion corresponds to the trajectory of an agent in time. Hence the modeling task now consists in construction of a formula in which satisfying assignments correspond to directed paths from $s_0^0(a_i)$ to $g^{t_M}(a_i)$ in the time expansion.

The time expansion is often further improved when used in MAPF so that unreachable nodes are removed which reduces the subsequent search effort done on top of the time expansion \cite{DBLP:conf/iccad/Bryant95,DBLP:journals/ai/SharonSGF13,DBLP:conf/ecai/SurynekFSB16}. A structure called {\em multi-value decision diagram} (MDD) is a subset of the time expansion as described above, that is, it is a directed graph $\mathit{MDD}_i=(V_i,E_i)$ for each agent $a_i$. A vertex $v^t$ is included in $\mathit{MDD}_i$ if is reachable with respect to the given makespan bound $t_M$. That is, $v^t$ is included if and only if $\mathit{Dist}(s_0(a_i),v) \leq t$ (agent $a_i$ has enough time to reach $v$ at time step $t$) and $\mathit{Dist}(v,g(a_i)) \leq t_M - t$ (agent $a_i$ can reach its goal in the remaining time) where $\mathit{Dist}(u,v)$ is the length of a shortest path between $u$ and $v$ in $G$.

\subsection{Hamiltonian Multi-value Decision Diagram}

The idea of $\mathit{MDD}$ can be adapted for MAPF$^{MG}$. We introduce a {\em Hamiltonian} $\mathit{MDD}$ for individual agents denoted $\mathit{MDD}^H_i$ which is almost identical to $\mathit{MDD}_i$ in the terms of time expansion of underlying $G$. However, the reachability of vertices at given time steps in $\mathit{MDD}^H_i$ is treated in a different way to reflect that each agent in MAPF$^{MG}$ has multiple goal vertices to visit.

A node $v^t$ is included in $\mathit{MDD}^H_i$ iff the following conditions hold:

\begin{enumerate}
\item $\mathit{Dist}(s_0(a_i),v) \leq t$ and
\item a Hamiltonian path $H_P=H_P(s_0(a_i), g(a_i) \cup \{v\})$ exists in $G$ such that $\mathit{Cost}(H_P) \leq t_M$.
\end{enumerate}

In other words, time $t$ must be enough to reach $v$ from the start and there must be a Hamiltonian path starting at $s_0(a_i)$ visiting both $v$ and all agent's goals of the cost not exceeding the number of levels of expansion $t_M$. Without proof let us summarize important property of $\mathit{MDD}^{H_P}$.

\begin{proposition}
A sequence of vertices visited by an agent $a_i$ within any solution to MAPF$^{MG}$ $\Theta$ with the sum-of-costs $SoC$ corresponds to a directed path in $\mathit{MDD}^{H_P}_i$.
\end{proposition}

Computing costs of Hamiltonian paths for every vertex in $G$ requires substantial computational effort as finding Hamiltonian path itself is an NP-hard problem \cite{DBLP:journals/siamcomp/GareyJT76} and also confirmed by our preliminary tests. Hence we introduce a relaxation of $\mathit{MDD}^{H_P}$ denoted $\mathit{MDD}^{T_S}$ in which the requirement of the existence of a Hamiltonian path is replaced with the requirement of existence of a {\bf spanning tree} $T_S(s_0(a_i),  g(a_i) \cup \{v\})$ of cost at most $t_M$. Computing a spanning tree can be done in polynomial time. A direct corollary of Proposition \ref{prop-Hamiltonian} is that using $\mathit{MDD}^{T_S}$ is sound for subsequent construction of the formula.

\begin{corollary}
A sequence of vertices visited by an agent $a_i$ within any solution to MAPF$^{MG}$ $\Theta$ with the sum-of-costs $SoC$ corresponds to a directed path in $\mathit{MDD}^{T_S}_i$.
\end{corollary}

\begin{figure}[t]
    \centering
    \includegraphics[trim={3.0cm 19cm 2.0cm 5.0cm},clip,width=0.55\textwidth]{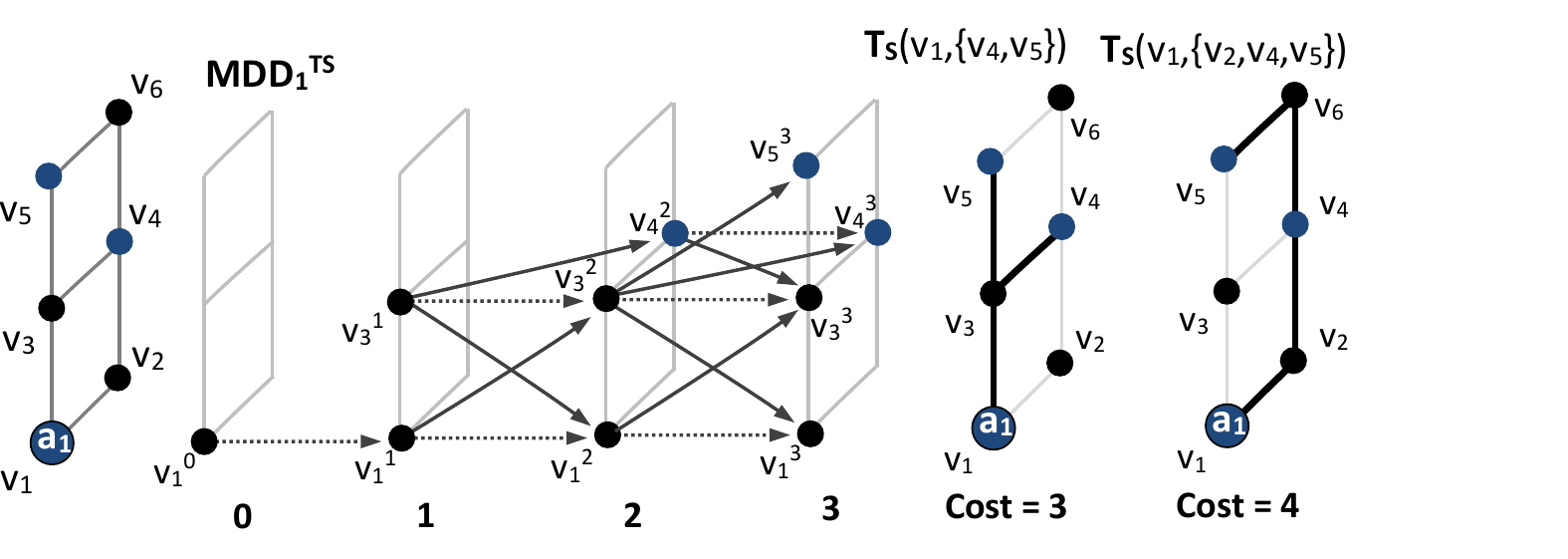}
    \vspace{-0.8cm}\caption{$\mathit{MDD}^{T_S}$ expanded for $t_M=3$. Spanning trees including various vertices are shown too. }
    \label{figure-MDD-HP}
\end{figure}

See figure \ref{figure-MDD-HP} for illustration how $\mathit{MDD}^{T_S}$ can be used to reduce the number of nodes in the time expansion. A spanning tree of cost 3 including vertices  $v_1$, $v_3$ and $v_5$ exists so these vertices reachable at time step 3 from $v_1$ are included but $v_2$ not because no spanning tree of cost less than 4 including  $v_1$, $v_2$, $v_3$ and $v_5$ exists in $G$ (similarly for $v_6$).

As solutions to MAPF$^{MG}$ are typically longer than those in the standard MAPF such pruning of the time expansion is cruicial to keep the resulting formula reasonably small.

\subsection{Incomplete Propositional Model}

We construct a formula $\mathcal{H(\mathit{SoC})}$ representing an {\em incomplete model} for MAPF$^{MG}$. Assume we have $\mathit{MDD}^{T_S}_i=(V_i,E_i)$ for agent $a_i$, then can derive the formula from $\mathit{MDD}^{T_S}_i$ as follows. A propositional variable $\mathcal{X}_v^t(a_j)$ is introduced for every vertex $v^t$ in $V_i$. The semantics of $\mathcal{X}_v^t(a_i)$ is that it is $\mathit{TRUE}$ if and only if agent $a_i$ resides in $v$ at time step $t$. Similarly we introduce $\mathcal{E}_{u,v}^t(a_i)$ for every directed edge $(u^t,v^{t+1})$ in $E_i$. Analogously the meaning of $\mathcal{E}_{u,v}^t(a_i)$ is: it is $\mathit{TRUE}$ if and only if agent $a_i$ traverses edge $\{u,v\}$ between time steps $t$ and $t+1$.

Constraints are added on top of these variables to encode the movement rules of MAPF$^{MG}$ that are exactly the standard MAPF rules such as that agents move across edges (do not jump) and preserve basic physical properties (do not disappear and do not spawn). The detailed list of constraints is given in \cite{DBLP:conf/ecai/SurynekFSB16} and \cite{DBLP:conf/ijcai/Surynek19}, for the sake of brevity we show here only illustrative examples:

\begin{equation}
   {  \mathcal{X}_u^t(a_i) \Rightarrow \bigvee_{(u^t,v^{t+1}) \in E_i}{\mathcal{E}^t_{u,v}(a_i),}
   }
   \label{eq:basic-start}
\end{equation}
\begin{equation}
   {  \sum_{v^{t+1}\:|\:(u^t,v^{t+1}) \in E_i }{\mathcal{E}_{u,v}^t{(a_i)} \leq 1}
   }
   \label{eq-2}
\end{equation}

These constraints state that if agent $a_i$ appears in vertex $u$ at time step $t$ then it has to leave through exactly one edge $(u^t,v^{t+1})$.

MAPF$^{MG}$ differs from MAPF in the treatment of encoding the goal condition. An agent $a_i$ in MAPF$^{MG}$ must visit each of its goal at least once. Hence we add for each $v \in g(a_i)$ the following constraint, that is we take all copies of $v$ in $\mathit{MDD}^{T_S}_i$ and require that at least one of the corresponding propositional variables must be $\mathit{TRUE}$:

\begin{equation}
   {  \sum_{v^{t}\:|\:v^t \in V_i }{\mathcal{X}_{v}^t{(a_i)} \geq 1}
   }
   \label{eq-goal}
\end{equation}

The incompleteness of the model as suggested in \cite{DBLP:conf/ijcai/Surynek19} consists in omitting certain constraint constraints. Specifically collision avoidance constraints are omitted. Hence instead of the equivalence between satisfiability of $\mathcal{H(\mathit{SoC})}$ and solvability of MAPF$^{MG}$ under given $\mathit{SoC}$ we only establish an {\bf implication} as follows:

\begin{definition}
  {\bf (incomplete propositional model).} Propositional formula $\mathcal{H(\mathit{SoC})}$ is an {\em incomplete propositional model} of MAPF$^{MG}$ 
  $\Theta$ if the following condition holds:
  \begin{center}
  $\mathcal{H}(\mathit{SoC}) \in \mathit{SAT}$ $\Leftarrow$ $\Theta$ has a solution of sum-of-costs $\mathit{SoC}$.
  \end{center}
  \label{definition-incomplete}
\end{definition}

\subsection{The Algorithm and Integrated Goal Ordering}
After obtaining a satisfiable formula we cannot immediately declare it to represent a valid solution but first the extracted candidate for MAPF$^{MG}$ solution must be verified for collisions. If threre are no conflicts then then we are finished and the candidate is a valid MAPF$^{MG}$ solution. If not then a collision avoidance constraint must be added to refine $\mathcal{H(\mathit{SoC})}$ and solving continues with the next iteration. Such a tight integration of lazy formula construction and its solving using the SAT solver is often used in the {\em satisfiability modulo theories} (SMT) paradigm \cite{DBLP:conf/fmcad/KatzBTRH16}.

Assume that a collision occurs at time step $t$ at vertex $v$ between agents $a_i$ and $a_j$ (denoted  $(a_i,a_j,v,t)$). Eliminating this collision is to forbid that $a_i$ or $a_j$ does not appear at $v$ at time step $t$ which naturally corresponds to a binary clause: $\neg \mathcal{X}_v^t(a_i) \vee \neg \mathcal{X}_v^t(a_j)$. Being aware of the construction of modern CDCL SAT solvers \cite{DBLP:journals/ijait/AudemardS18} we can see that this is quite fortunate as such short clauses promotes {\em Boolean constraint propagation} (resolution, unit propagation) that significantly reduces the search effort.

The resulting algorithm called SMT-HCBS is described using pseucode as Algorithm \ref{alg-SMTHCBS}. It is a modification of existing SMT-CBS, the major difference is the use of specific propositional encoding for MAPF$^{MG}$ (line 11) as described above. The algorithm tests the existence of a solution of the input instance $\Theta$ for increasing sum-of-costs until a positive answer is obtained (lines 5-9). The lower bound for sum-of-costs is obtained as the sum-of-costs of individual Hamiltonian paths for agents (lines 3-4).

The advantage against {\em complete models} \cite{DBLP:conf/ecai/SurynekFSB16} is that a valid solution can be obtained before the problem is fully specified in terms of constraints which often leads to faster solving process. Observe that unsatisfiable formula in the case of incomplete model according to Definition \ref{definition-incomplete} means that the input instance $\Theta$ is not solvable under the given cost $\mathit{SoC}$, leading to its incrementing and the next iteration of the algorithm.

\begin{algorithm}[h]
\begin{footnotesize}
\SetKwBlock{NRICL}{SMT-HCBS ($\Theta = (G=(V,E),A,s_0,g)$)}{end} \NRICL{
    $conflicts \gets \emptyset$\\
    $circuits \gets$ $\{circuit^*(a_i)$ a shortest Hamiltonian path from $s_0(a_i)$ covering $g(a_i)\;|\; i = 1,2,...,k\}$ \\
    $\mathit{SoC} \gets \sum_{i=1}^k{\mathit{Cost}(circuits(a_i))}$ \\    
    \While {$TRUE$}{
         $(circuits,conflicts) \gets$ SMT-HCBS-Fixed($conflicts,\mathit{SoC},\Theta$)\\
        \If {$circuits \neq$ UNSAT}{
        	\Return $circuits$\\
        }
        $\mathit{SoC} \gets \mathit{SoC} + 1$ \\
    }
}  

\SetKwBlock{NRICL}{SMT-HCBS-Fixed($conflicts,\mathit{Soc},\Theta$)}{end} \NRICL{
	    $\mathcal{H}(\mathit{SoC}) \gets$ encode-Hamiltonian$(conflicts,\mathit{SoC},\Theta)$\\
	    \While {$TRUE$}{
	        $assignment \gets$ consult-SAT-Solver$(\mathcal{H}(\mathit{SoC}))$\\
	        \If {$assignment \neq UNSAT$}{
	            $circuits \gets$ extract-Solution$(assignment)$\\
	            $collisions \gets$ validate($circuits$)\\
                   \If {$collisions = \emptyset$}{
                      \Return $(circuits,conflicts)$\\
                   }
                   \For{each $(a_i,a_j,v,t) \in collisions$}{
                      $\mathcal{H}(\xi) \gets \mathcal{H}(\xi) \cup \{\neg \mathcal{X}_v^t(a_i) \vee \neg \mathcal{X}_v^t(a_j)$\}\\
                      $conflicts \gets conflicts \cup \{[(a_i,v,t),(a_j,v,t)]\}$
                   }
               }
               \Return {(UNSAT,$conflicts$)}\\
          }
}

\caption{SMT-based  MAPF$^{MG}$ solver} \label{alg-SMTHCBS}
\end{footnotesize}
\end{algorithm}

Let us note that both goal vertex ordering and conflict resolution are integrated at the same conceptual level - both problems are encoded in the target formula are decided by the SAT solver.

\section{Experimental Evaluation}

We evaluated HCBS and SMT-HCBS on standard benchmarks from \texttt{movingai.com} \cite{DBLP:journals/tciaig/Sturtevant12}. Representative part of results is presented in this section.

\subsection{Benchmarks and Setup}

HCBS and SMT-HCBS were implemented in C++. The SMT-HCBS solver is built on top of the Glucose 3.0 SAT solver \cite{DBLP:journals/ijait/AudemardS18} that ranks among the high performing SAT solvers according to recent SAT solver competitions \cite{DBLP:conf/aaai/BalyoHJ17}.

The experimental evaluation has been done on diverse instances consisting of 4-connected {\em grid} maps ranging in sizes from small to large. Random MAPF scenarios from \texttt{movingai.com} are used to generate MAPF$^{MG}$ instances. To obtain instances of various difficulties we varied the number of agents while the number of goal vertices per agent was set as constant. As defined in the benchmark set, 25 different instances are generated per number of agents.

Starting positions of agents are taken directly from the scenario. Since only one goal is defined per agent in a MAPF scenario the set of goals for an agent is generated by making a given number of random picks among goal positions of all agents in the scenario. This results in MAPF$^{MG}$ instances whose goal vertices are equally distributed across the map.

The tests we focused on comparing HCBS and SMT-HCBS in the terms of success rate and runtime. All experiments were run on system consisting of 200 Xeon 2.8 GHz cores, 1TB RAM, running Ubuntu Linux 18. \footnote{To enable reproducibility of presented results we provide complete source code of our solvers and detailed experimental data on author's web: \texttt{http://users.fit.cvut.cz/$\sim$surynpav/research}. and git repository: \texttt{https://github.com/surynek}.}
The success rate is the ratio of the number of instances out of 25 per number of agents that the solver managed to solver under the time limit of 5 minutes.

\begin{figure}[t]
    \centering
    \includegraphics[trim={2.5cm 2.5cm 2cm 2.5cm},clip,width=0.5\textwidth]{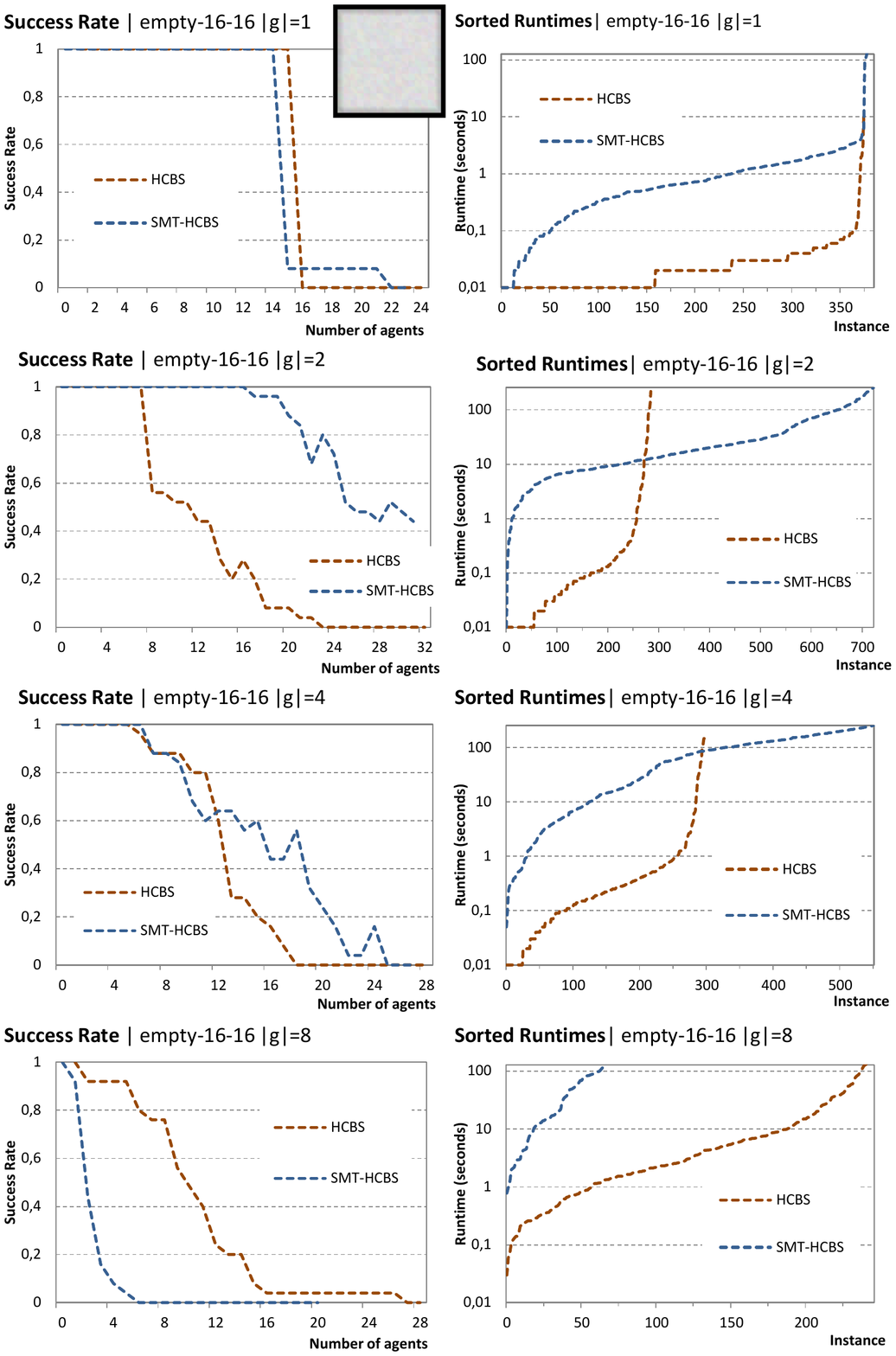}
    \vspace{-1.0cm}\caption{Success rate and runtime comparison on small-sized maps.}
    \label{expr-small}
\end{figure}

\subsection{Runtime Results}

We divided the tests into three categories with respect to the size of maps. Results for {\bf small} instances derived from the \texttt{empty-16-16} map are shown in Figure \ref{expr-small}. Three different cases with the number of goal vertices per agent: 1, 2, 4, and 8 are tested. Let us note that one goal vertex correspond to the standard MAPF.

We can observe that for 1, 2 and 4 goals per agent SMT-HCBS dominates. This is an expectable result since similar behaviour can be observed for SMT-CBS vs. CBS for the standard MAPF \cite{DBLP:conf/ijcai/Surynek19}. However the clearly visible trend it that HCBS scales better for increasing number of goals which eventually in reversed situation with 8 goals where HCBS performs better.

\begin{figure}[t]
    \centering
    \includegraphics[trim={2.5cm 14.0cm 2cm 2.5cm},clip,width=0.5\textwidth]{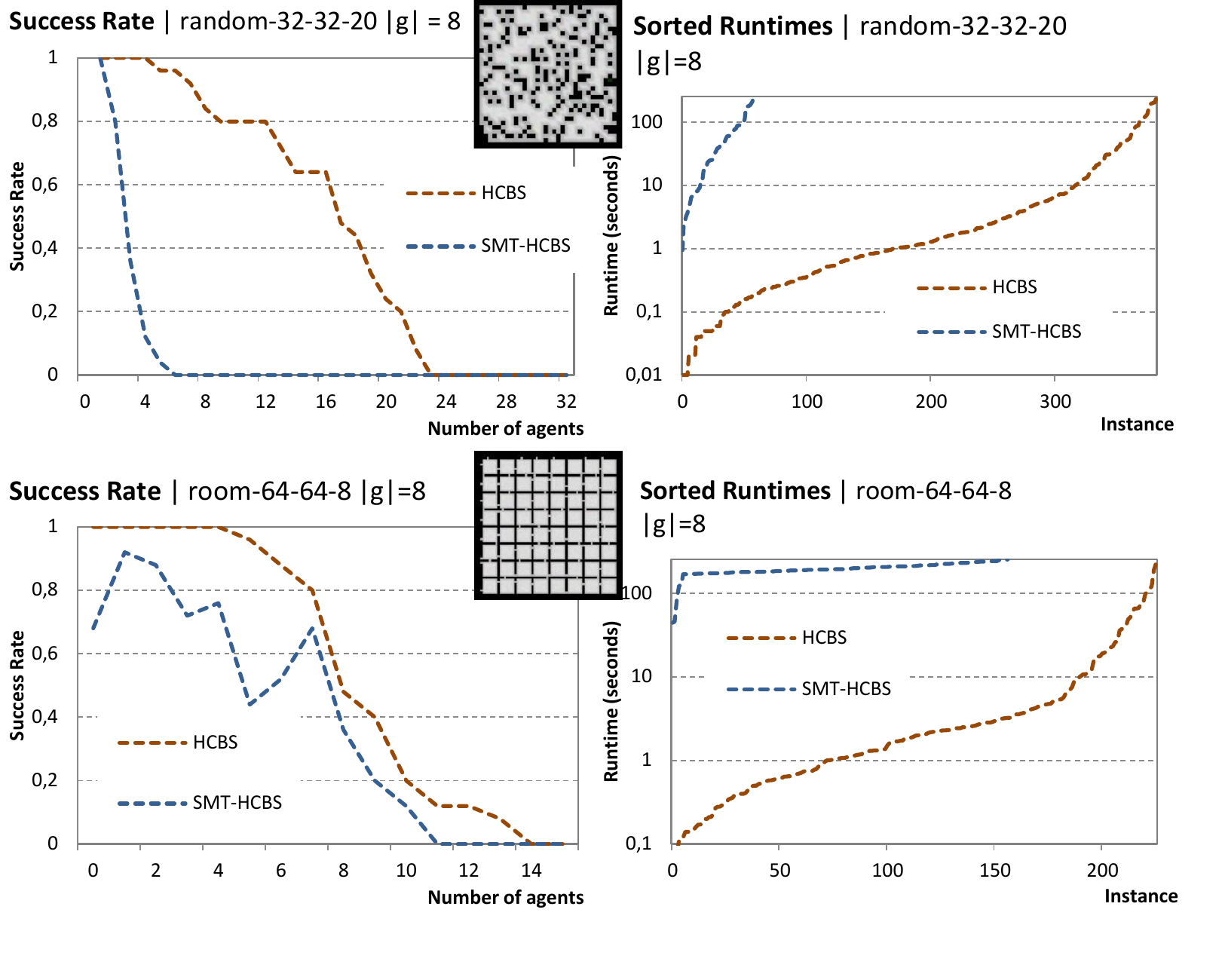}
    \vspace{-1.0cm}\caption{Success rate and runtime comparison on medium-sized maps.}
    \label{expr-medium}
\end{figure}

Results for {\bf medium-sized} instances shown in Figure \ref{expr-medium} where 8 goals per agent were used indicate that HCBS perform significantly better than SMT-HCBS. Only relatively closer performance to HCBS was achieved by SMT-HCBS on instances derived from the \texttt{room-64-64-8} map.

\begin{figure}[t]
    \centering
    \includegraphics[trim={2.5cm 8.5cm 2cm 2.5cm},clip,width=0.5\textwidth]{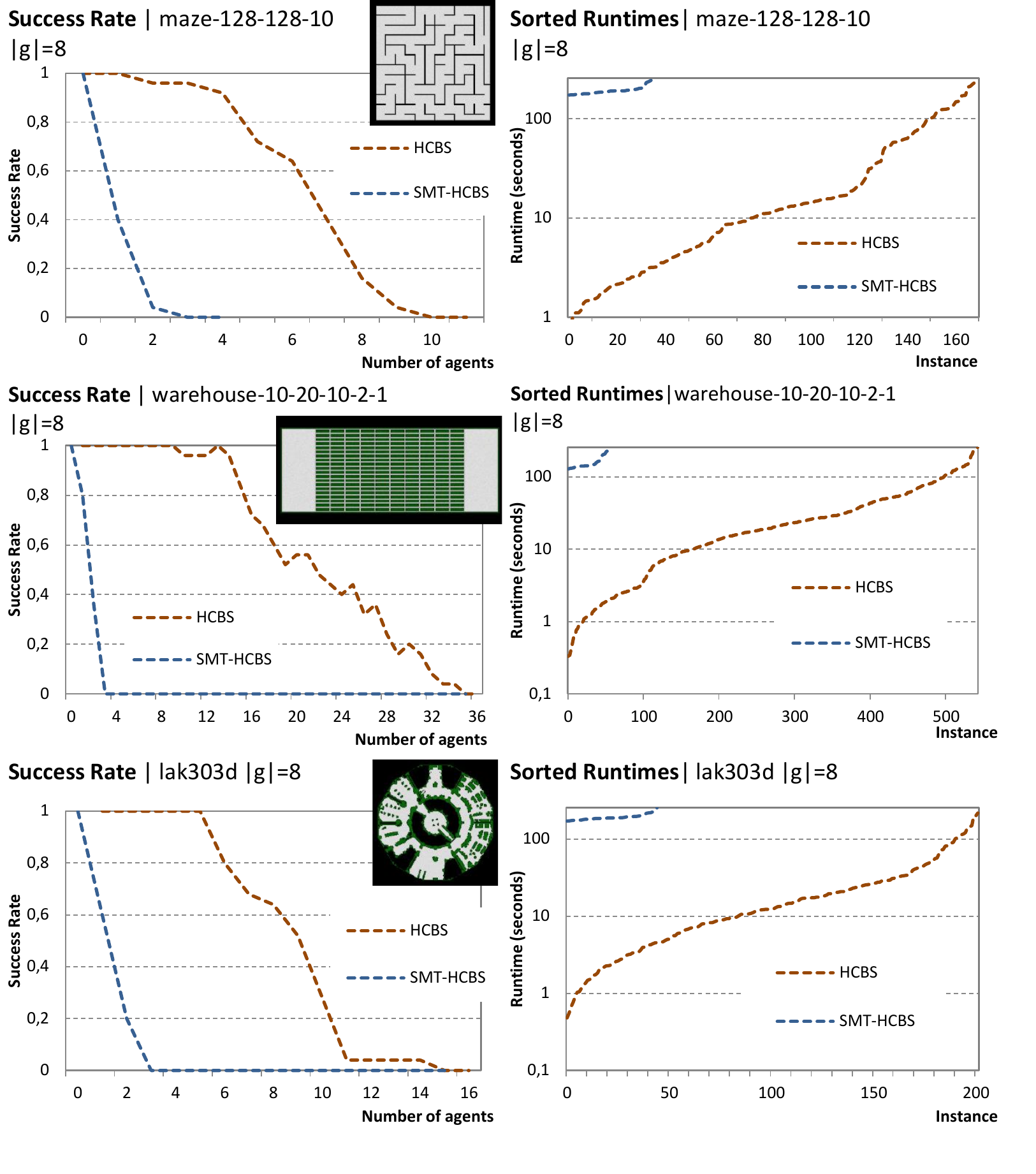}
    \vspace{-1.0cm}\caption{Success rate and runtime comparison on large-sized maps.}
    \label{expr-large}
\end{figure}

Eventually results for {\bf large} instances shown in Figure \ref{expr-large} indicate clear dominance of HCBS over SMT-HCBS.

\subsection{Explanation of Results}
There are multiple factors explaining the observed results. SMT-HCBS often needs to iterate through many unsatifiable sum-of-costs before the optimal sum-of-cost is met on instances with greater number of goal vertices. This is due to the fact that agents interact more in such cases which results in a greater difference between the true optimal sum-of-costs and its lower bound estimation. Moreover node pruning in spanning tree MDDs is less efficient for greater number of goals since the cost of spanning tree deviates more from the true Hamiltonian path cost.

On the other hand, the effect of domain specific heuristics is more direct in HCBS which can more effectively determine the optimal ordering of goal vertices for individual agents - the choice of the next vertex can be immediately assessed by the heuristic. Similar effect cannot be easily achieved in SMT-HCBS since during MDD generation phase we need to represent all possible goal vertex orderings and the SAT solver itself has no domain specific information.

\section{Related Work}

MAPF and its variants have been intensively studied recently. The most closely related problem is {\em multi-agent pickup and delivery} (MAPD) \cite{DBLP:conf/atal/0001LKK17,DBLP:conf/atal/Liu0LK19}, defining a set of tasks $T=\{t_1,t_2,...t_m\}$ where each task $t_j=(p_j,d_j)$ is characterized by a pickup location $p_j \in V$ and a delivery location $d_j \in V$. Agents can freely select tasks to fulfill. Having freedom in choice of tasks and the ordering of their fulfilling makes the problem very hard. The contemporary solving approaches for MAPD first assign tasks to agents and followed by determining the ordering of tasks per agent ignoring collisions. Then collision free paths are planned according the task ordering. As there is no feedback between the phases the resulting plan is sub-optimal.

Another related problem is represented by order picking problem (OPP) \cite{DBLP:journals/cor/PansartCC18} which is a variant of traveling salesman problem in a rectangular warehouse. Various integer programming and flow-based formulations of OPP have been studied. The important difference from MAPF$^{MG}$ is that it is typically regarded as a single agent problem and hence collisions between agents are not considered in OPP.

A generalization of MAPF where an agent is assigned multiple goal vertices and is successful if it reaches any of its goals is suggested in \cite{DBLP:conf/sara/Surynek13}. Giving agents multiple options to finish their path makes the problem easier.

\section{Conclusion}

We introduced multi-goal multi agent path finding (MAPF$^{MG}$) in this work. MAPF$^{MG}$ generalizes MAPF by assigning each agent a set of goal vertices instead of one goal. The task is to visit each of agent's goal vertices at least once while we aim on finding sum-of-costs optimal solutions. Two algorithms generating sum-of-costs optimal solutions are suggested. A search-based algorithm called Hamiltonian CBS (HCBS) derived from CBS and a compilation-based approach SMT-HCBS that reduces MAPF$^{MG}$ to propositional logic and solves the problem in the target formalism by the SAT solver.

HCBS introduces three level search in which conflict resolution is done at the high level and goal vertex ordering and path planning are done at the low level. The key technique is decoupling vertex ordering from collision-free path planning. CBS provides greater room for integrating powerful heuristics that we made use of when adapting it for MAPF$^{MG}$.

On the other hand SMT paradigm provides more powerful search but integration of domain specific heuristics is more difficult. We tried to increase the informedness of the SMT-based algorithm SMT-HCBS through the concept of Hamiltonian and spanning tree MDD which eventually led to significant improvements so that SMT-based approach is able to solve some of the standard benchmark problems. However decoupling of goal vertex ordering from conflict resolution turned out to be crucial factor behind the better performance of search-based HCBS. The worse support of domain specific heuristics shows limitations of the SMT-based approach to MAPF$^{MG}$ and MAPF generally.

For future work we plan investigate possibility of generating the formulae in the SMT-based approach lazily not only in the phase of conflict resolution but also in the phase of MDD generation. One possible future direction is also to address MAPF$^{MG}$ with more active use of the SAT solver in DPLL(MAPF$^{MG}$) framework \cite{DBLP:conf/aiia/Surynek19}.

\begin{quote}
\begin{small}
\bibliography{bibfile}
\end{small}
\end{quote}

\bigskip
\end{document}